\documentclass{ecai}  

\usepackage{graphicx}
\usepackage{latexsym}

\paperid{1202}        

\usepackage{comment}
\usepackage{caption}
\usepackage{subcaption}
\usepackage{tabularx}
\usepackage{threeparttable}
\usepackage{multirow}
\usepackage{xcolor}
\usepackage[acronym]{glossaries}
\usepackage{eqparbox}
\usepackage{algorithmic}
\usepackage{algorithm}
\renewcommand\algorithmiccomment[1]{
	\hfill\ $\triangleright$ \ \eqparbox {COMMENT}\small{{#1}}}
\newacronym{ONEES}{\textit{EF-OnTL}}{\textit{Expert-Free Online Transfer Learning}} 
\newacronym{sarnd}{sars-RND}{\textit{State Action Reward Next-State Random Network Distillation}}

\newacronym{rdc}{\textit{rnd $\Delta$-conf}}{\textit{Random transfer from Delta Confidence with threshold}}
\newacronym{hdc}{\textit{high $\Delta$-conf}}{\textit{transfer Higher Delta Confidence}}
\newacronym{lec}{\textit{loss $\&$ conf}}{\textit{higher Loss and Confidence mixed}}
\newacronym{pp}{MT-PP}{Multi-Team Predator-Prey}
\newacronym{u}{$\overline{U}$}{\textit{Average Uncertainty}}
\newacronym{bp}{BP}{\textit{Best Performance}}
\newacronym{exp_adv}{\textit{RCMP}}{\textit{Requesting Confidence-Moderated Policy advice}~\cite{da2020uncertainty}}
\newacronym{noexp_adv}{\textit{OCMAS}}{\textit{Online Confidence-Moderated Advice Sharing}}
\newacronym{HFO}{HFO}{Half Field Offense}
\usepackage{pifont}
\usepackage{xcolor}
\newcommand{\cmark}{{\ding{51}}}
\newcommand{\xmark}{{\ding{55}}}

\begin{document}
	
	\begin{frontmatter}
		
		\title{Expert-Free Online Transfer Learning in Multi-Agent Reinforcement Learning}
		
		\author[A]{\fnms{Alberto}~\snm{Castagna}\thanks{Corresponding Author. Email: acastagn@tcd.ie.}}
		\author[A]{\fnms{Ivana}~\snm{Dusparic}}
		
		\address[A]{School of Computer Science and Statistics, Trinity College Dublin, Dublin, Ireland}

		\begin{abstract}
			Transfer learning in Reinforcement Learning~(RL) has been widely studied to overcome training challenges in Deep-RL, i.e., exploration cost, data availability and convergence time, by bootstrapping external knowledge to enhance learning phase. 
			While this overcomes the training issues on a novice agent, a good understanding of the task by the expert agent is required for such a transfer to be effective. 
			
			As an alternative, in this paper we propose \acrfull{ONEES}, an algorithm that enables expert-free real-time dynamic transfer learning in multi-agent system. No dedicated expert agent exists, and transfer source agent and knowledge to be transferred are dynamically selected at each transfer step based on agents' performance and level of uncertainty. To improve uncertainty estimation, we also propose~\acrfull{sarnd}, an extension of RND that estimates uncertainty from RL agent-environment interaction.
			
			We demonstrate \acrshort{ONEES} effectiveness against a no-transfer scenario and state-of-the-art advice-based baselines, with and without expert agents, in three benchmark tasks: Cart-Pole, a grid-based \acrfull{pp} and \acrfull{HFO}.

			Our results show that \acrshort{ONEES} achieves overall comparable performance to that of advice-based approaches, while not requiring expert agents, external input,  nor threshold tuning. \acrshort{ONEES} outperforms no-transfer with an improvement related to the complexity of the task addressed.
		\end{abstract}
		
	\end{frontmatter}
	
	\section{Introduction}
	Transfer learning~(TL) in reinforcement learning~(RL) has been introduced to address two main shortcomings of RL: data availability, as acquiring sufficient interaction samples can be prohibitive due to large state and action space, and 
	lowering the exploration cost due to the partial observability,
	sparse feedback and safety concerns that may incur in real world environments~\cite{dulac2021challenges}.

	Most TL in RL solutions utilize an expert, either agent or human, which supervises a novice agent through the Teacher-Student framework~\cite{torrey2013teaching}. This can be exploited in two ways, by supporting other agents during exploration, i.e., \cite{liu2022, nair2018overcoming,subramanian2022multi,zhu2020learning}, or by training a novice agent to emulate an expert behaviour, i.e.,~\cite{hester2017learning,taylor2018improving}. While this solves the issue for "student" agents, by lowering the amount of required training data and time for a specific task, it does not solve the broader problem, as an expert agent is required for each task. Furthermore, there may be task where an expert is unavailable to support student agents or does not yet exist. 
	
	This paper addresses the above RL limitations by introducing \acrfull{ONEES}. \acrshort{ONEES} removes the need for a fixed expert in the system.
	Instead, it takes advantage of different knowledge gained in different parts of a multi-agent system. At each step, a current source agent is dynamically selected based on system performance and/or epistemic confidence. 
	The source agent's experience is partially transferred to target agents. Roles are assigned at each transfer step, while  knowledge shared are agent-environment transitions~$(s^t, a^t, r^t, s^{t+1})$ labelled with agent's epistemic uncertainty~($u^t$).

	To estimate uncertainty, we propose an extension of Random Network Distillation~RND~\cite{burda2018exploration}: \acrfull{sarnd}. 
	\acrshort{sarnd} estimates agent's epistemic uncertainty from the full agent-environment interaction.
	Thus, while RND approximates a state-visit counter, \acrshort{sarnd} provides a more fine-grained estimation based on the action taken and on its outcome.

	Source selection relies on two criteria, \acrfull{u} and \acrfull{bp}. \acrshort{u} evaluates uncertainties over collected samples while \acrshort{bp} analyses agent's performance in a given interval. Once transfer roles are defined for the time step, target agents receive a personalised batch of experience, selected based on sender-receiver uncertainties and expected surprise. Expected surprise is estimated by computing the expected loss on target's side~\cite{white2014surprise}.
	We compare our proposed method against three baselines: (i) a \textit{no-transfer} scenario; (ii) baseline in which, instead of transferring experience, an advice is transferred, i.e. a recommended action (in order to evaluate the impact of sharing advice versus experience with~\acrfull{noexp_adv}), as work in~\cite{ilhan2019teaching} 
	 and (iii) state-of-the-art advice-based approach \acrfull{exp_adv}, in which previously trained agents provide action-advice on demand, when an agent is in a state with high uncertainty. 
	
	We evaluate \acrshort{ONEES} in three environments, (1) Cart-Pole, (2) \acrfull{pp} and (3) \acrfull{HFO}~\cite{ALA16-hausknecht}.

	Therefore, contribution of this paper is threefold:
	\begin{itemize}
		\item We propose \acrshort{ONEES}, an algorithm that enables online transfer learning by experience-sharing without a need for an expert agent and is applicable to a range of RL-methods;
		\item We propose \acrshort{sarnd}, an extension of RND, that  estimates uncertainty based on a full RL environment transition tuple~$(s^t, a^t, r^t, s^{t+1})$;
		\item We assess several criteria to enable dynamic selection for source of transfer and shared knowledge.
		As result, at each step, each target agent receives a tailored batch of experience, specific to its current belief.
	\end{itemize}

	The rest of this paper is organized as follows. Related work on agent-to-agent TL and overview of uncertainty estimators is presented in ~Section~\ref{sec:related_work}.
	\acrshort{ONEES}, \acrshort{sarnd} and proposed criteria are detailed in Section~\ref{sec:contributions}. Simulation setup is in Section~\ref{sec:simulations} while evaluation results are presented in Section~\ref{sec:results}. Lastly, Section~\ref{sec:conclusion} discusses the framework's limitations and future work directions.
	\section{Related Work}\label{sec:related_work}
	
	This section provides an overview of the existing work on {agent-to-agent} TL and methods used to approximate agent's epistemic uncertainty.

	\subsection{Transfer Learning in RL}
	Most of agent-to-agent TL work is based on the {teacher-student} framework~\cite{torrey2013teaching} where the teacher is an expert and the student is a novice agent. The student can request the teacher's supervision, and due to resource constraints, their interaction is limited by a budget.

	Preferred form of advice is action-based, i.e., \cite{da2017simultaneously,da2020uncertainty,  ilhan2019teaching, liu2022, norouzi2021experience, subramanian2022multi,taylor2019parallel, taylor2014reinforcement, torrey2013teaching, zhu2020learning}. This lowers the overall exploration cost for the novice by asking for an action to follow in certain states. 
	Other form of advice is Q-values~\cite{liang2020parallel,liu2022,zhu2021q} used to influence the action-selection process of a target agent.
	Finally, advice can be provided as policy to be followed for a certain number of steps~\cite{yang2020learning} or a batch of RL experience, similar to a demonstration, as in~\cite{Castagna2022MultiAgentTL,cruz2017pre,gabriel2019pre,nair2018overcoming,wang2017improving}.
	
	Despite the good result shown by the teacher-student framework, roles are kept fixed and transfer impact is limited by the teacher knowledge. Therefore, insufficiently skilled expert might worsen novice's performance. 
	To overcome this limitation, \cite{norouzi2021experience, yang2020learning} introduce a real-time training of a centralised super-entity based exclusively on a subset of collected demonstration. This new entity provides on-demand advice when needed. While that improves advice quality over time, introducing a new super agent results in an additional cost to gather the experiences and to train the underneath model.
	\cite{da2017simultaneously, ilhan2019teaching, liang2020parallel, taylor2019parallel,zhu2021q} improved the base teacher-student framework by relying on confidence-based and importance-based methods to dynamically select one or more agents as teachers. In these models, an agent can ask for and provide advice simultaneously. 

	Baselines used in this paper are based on the above state-of-the-art methods: \acrshort{noexp_adv} is based on~\cite{ilhan2019teaching} and \acrshort{exp_adv} on~\cite{da2020uncertainty}. In~\cite{ilhan2019teaching}, an uncertain agent asks for advice and more confident agents recommend an action to follow. Confidence is measured through \acrshort{sarnd} and advice is given only when the advice seeker is the most uncertain agent across the team. 
	Final action is taken by majority voting. In \acrshort{exp_adv}~\cite{da2020uncertainty}, an agent asks for advice during exploration based on its epistemic uncertainty. Advice, in form of an action, is given by a fixed expert-demonstrator. In our implementation, we replaced the single advisor with a jury of multiple trained agents to ward off any bias that might arise from using a single trained agent as teacher and action is selected by majority~voting.
	
	Table~\ref{tab:rel_work_summary} summarises relevant recent work to this research. For each method, we indicate whether a prior external expert is needed or not. Note that we do not distinguish teacher expertise, i.e., optimal, sub-optimal, etc. When expert is not required, we specify whether the algorithm enables dynamic roles or not. Lastly, we report type of advice transferred and the frequency of transfer. The latter does not take into account when nor how often the advice is used.
	\begin{table}
		\renewcommand*{\arraystretch}{1.1}
		\caption{\label{tab:rel_work_summary} Summary of related work.}
		\centering
		
		\begin{threeparttable}
			\begin{tabular}{ccccc}
				\hline
				\multirow{2}{*}{\textbf{Reference}}&\textbf{Expert} &\textbf{Dynamic} & \textbf{Transferred}&\textbf{Transfer}\\
				\textbf{}&\textbf{required} &\textbf{roles} & \textbf{advice}&\textbf{Frequency}\\\hline
				\cite{da2020uncertainty,subramanian2022multi,taylor2014reinforcement} & \multirow{2}{*}{\cmark} & \multirow{2}{*}{\xmark} & \multirow{2}{*}{action}&\multirow{2}{*}{dynamic} \\
				\cite{torrey2013teaching,zhu2020learning}&&&&\\
				\cite{da2017simultaneously,ilhan2019teaching,liang2020parallel} & \xmark & \cmark & action & dynamic \\
				\cite{taylor2019parallel,zhu2021q} & \xmark & \cmark & Q-values & dynamic\\
				\cite{norouzi2021experience} & \xmark & \xmark & action & dynamic\\
				\cite{yang2020learning} & \xmark & \xmark & policy & dynamic\\
				\cite{ Castagna2022MultiAgentTL,cruz2017pre,gabriel2019pre} & \multirow{2}{*}{\cmark} & \multirow{2}{*}{\xmark} & \multirow{2}{*}{RL-tuple} & \multirow{2}{*}{fixed}\\
				\cite{nair2018overcoming,wang2017improving} &&&&\\
				\multirow{2}{*}{\cite{liu2022}} & \multirow{2}{*}{\cmark} & \multirow{2}{*}{\xmark} & action $\&$ & \multirow{2}{*}{fixed}\\
				&&& Q-values &\\
				
				\textbf{\acrshort{ONEES}} & \textbf{\xmark} & \textbf{\cmark} & \textbf{RL-tuple} & \textbf{fixed} \\\hline
				
			\end{tabular}
		\end{threeparttable}
	\end{table}

	Our work differs from the above as it removes the need for presence of an expert agent by dynamically selecting a learning agent as a teacher, to exploit different knowledge gained in different parts of a multi-agent system. Furthermore, while most of the related work continuously override target policy by advising actions, Q-values and policy, \acrshort{ONEES} combines local and received knowledge by transferring, less frequently, an experience batch specifically selected based on a gap between source and target beliefs. Lastly, \acrshort{ONEES} does not require the tuning of additional parameters, i.e., uncertainty threshold, to improve the quality of transfer.

	\subsection{Uncertainty Estimation in TL}\label{ss:rw_uncertainty}
	In RL there are two types of uncertainties, \textit{aleatoric} and \textit{epistemic}.
	\textit{Aleatoric} comes from the environment and is generated by~stochasticity in observation, reward and actions.
	\textit{Epistemic} uncertainty comes from the learning model and indicates whether the agent has adequately explored a certain state.
	Most recent TL frameworks rely on epistemic uncertainty to determine if an agent requires guidance.
	
	A possible approach to approximating agent's epistemic uncertainty in a particular task is to determine the frequency of visits to each state. 
	In fact, \cite{da2017simultaneously, norouzi2021experience,taylor2019parallel} estimate uncertainty by relying on function defined over state visits counter. Similarly,~\cite{zhu2021q} relies on number of visits over a state-action pair.	However, state space might be continuous or very large making the counting unfeasible. Thus, state visit counts could be approximated by RND~\cite{burda2018exploration}. RND has been proposed to encourage exploration within agents but has already been exploited as uncertainty estimator for TL in RL in~\cite{Castagna2022MultiAgentTL, ilhan2019teaching}. RND consists of two networks, a target with an unoptimised and randomly initialised parameters and a main predictor network. Throughout time, the former is distilled within the latter and uncertainty is defined as prediction error between the two outputs.

	Other sophisticated models can be used to estimate the epistemic uncertainty. \cite{wang2017improving} proposes uncertainty estimation through neural network, decision tree and Gaussian process. Although, neural networks seem to be preferred overall as are also used in~\cite{da2020uncertainty}, where the agent's learning model is expanded by replacing control-layer with an ensemble to estimate agent's uncertainty. However, despite the different underlying technique used to estimate uncertainty, all the presented methods rely uniquely on visited state.
	
	To overcome this limitation, this paper introduces~\acrshort{sarnd} as an extension of RND. \acrshort{sarnd} computes epistemic uncertainty over a full RL interaction~$(s^t, a^t, r^t, s^{t+1})$.

	\section{Expert-Free Online Transfer Learning}\label{sec:contributions}
	
	This section introduces the three main contributions of this research: 
	(1) \acrshort{sarnd}:~Section~\ref{ss:sarnd} addresses the uncertainty estimator model used within this work;
	(2) \acrshort{ONEES}:~Section~\ref{ss:TL} provides the details of the core of our research work towards an autonomous framework for online  transfer learning;
	(3) Selection criteria: Section~\ref{ss:SS_metrics} presents two criteria used to dynamically select transfer source and Section~\ref{ss:TM_metrics}~introduces criteria used to identify RL experience worth to be shared, by evaluating source and transfer beliefs. These metrics enable the batch to be personalised based on specific target knowledge shortcomings.
	
	\subsection{\acrshort{sarnd}}\label{ss:sarnd}
	We propose \acrfull{sarnd} as an extension of RND to improve the estimation of epistemic uncertainty. When using RND as an uncertainty estimator, agent might lose important information that should be taken into account when computing uncertainty. For instance, in a sparse-reward environment and/or continuous control space, i.e, \acrshort{HFO} and our \acrlong{pp} implementation described at Section~\ref{sec:simulations}, by not considering other details, such as action, agent might become confident within a state where the goal fulfilment is close but not yet achieved.
	
	To overcome this potential limitation we propose~\acrshort{sarnd}, an uncertainty estimator that takes into account a full RL interaction~$(s^t, a^t,r^t, s^{t+1})$, rather than just state, to compute epistemic uncertainty of an agent at a specific time.

	\subsection{Transfer Framework}\label{ss:TL}
	
	This section outlines the core contribution of this work \acrfull{ONEES}. \acrshort{ONEES} is a novel online transfer learning algorithm that overcomes the need of an unique expert agent by dynamically selecting a temporary expert within each transfer iteration. Selected agent is used as source of transfer and some of its collected experience is made available to others. Target agent can then filter and sample a batch of experience to be integrated into its learning process and finally update its belief.
	Transferred batch contains five-elements tuples, $(s^t, a^t,r^t,s^{t+1},u^t)$, source agent's RL interaction at time $t$ and $u_t$, an additional value that identifies source's uncertainty over the RL interaction updated at time~$t$  of~visit.

	\acrshort{ONEES} is independent of an underlying RL algorithm used Thus, it can be exploited on a range of RL methods, both tabular and neural network-based ones.

	To define \acrshort{ONEES} framework, we specify the following: 
	\begin{itemize}
		\item $N$ $-$  number of RL-based agents available during a simulation; 
		\item a set of \textit{Agents}~$A = \{A_1,\dots,A_N\}$;
		\item a set of \textit{Learning Processes} $LP = \{LP_1,\dots,LP_N\}$ with the following relation $f_1= {(A_i,LP_j)\in A \times LP \leftrightarrow i=j }$;
		\item a set of \textit{Uncertainty Estimators} $UE = \{UE_1,\dots,UE_N\}$ and relation $f_2= {(A_i,UE_j)\in A \times UE \leftrightarrow i=j  }$;
		\item a set of \textit{Transfer Buffers} $TB =\{ TB_1,\dots,TB_N\}$ related to \textit{agents} by $f_3= {(A_i,TB_j)\in A \times TB \leftrightarrow i=j  }$;
		\item $B$ $-$ constrained \textit{Transfer Budget} to be used within a single transfer interaction;
		\item \textit{TF} $-$ \textit{Transfer Frequency} defined as number of episodes that occur between two consecutive transfer steps;
		\item \textit{SS} $-$ \textit{Source Selection} technique used to select source of transfer;
		\item \textit{TM} $-$ \textit{Transfer Methods} used to filter relevant knowledge to be transferred.
	\end{itemize}

	\begin{algorithm}[h!]
		\small
		\caption{\acrlong{ONEES}}
		\label{alg:online_trasfer}
		\begin{algorithmic}[1]
			\STATE Given: \textit{A, LP, UE, TB, B, TF, SS, TM, $f_1$, $f_2$, $f_3$}
			\FOR{\textit{ep} in \textit{Episodes}} 
			\FORALL[follow normal policy]{$A_i \in A$}
			\STATE get state $s_i^t$ for $A_i$ at time $t$
			\STATE sample an action $a_i^t$ based on $LP_i$
			\STATE perform a step and observe $o_i^t = (s_i^t,a_i^t,r_i^t,s_i^{t+1})$
			\STATE  estimate uncertainty  $u_i^t$ = $UE_i(o_i^t)$ 
			\STATE push ($o_i^t, u_i^t$) to $TB_i$ \algorithmiccomment{FIFO queue}
			\STATE optimize $UE_i$ on $o_i^t$
			\IF{time to update $LP_i$} 
			\STATE optimize $LP_i$ 
			\ENDIF
			\ENDFOR
			\IF[start transfer-step]{\textit{ep} \% $TF$ is $0$} 
			\STATE select source agent $A_s$ by $SS$ 
			\FORALL[transfer from $A_s$ to $A_t$]{$A_t \in (A \setminus A_s)$} 
			\STATE apply $TM$ over $TB_s$ and sample $B$ tuples
			\STATE update $LP_t$ with the sampled tuples
			\ENDFOR
			\ENDIF
			\ENDFOR
			
		\end{algorithmic}
	\end{algorithm}

	Algorithm \ref{alg:online_trasfer} introduces high level procedure followed by agents for sharing experience one to another throughout their simultaneous exploration processes.
	
	First, line~$1$ defines \acrshort{ONEES} parameters.
	Then, at lines~$3-6$ agents retrieve observation from an environment, sample an action based on their learning process and finally take a step. Afterward, each agent~$A_i$ estimates its epistemic uncertainty $u_i^t$ over the interaction~$(s_i^t,a_i^t,r_i^t,s_i^{t+1})$, updates its $UE_i$ model based on new sampled evidence and publishes its uncertainty-labelled collected tuple to the associated transfer buffer $TB_i$~($7-9$). When $TB_i$ is at full capacity, new labelled interactions replace the oldest tuples following a FIFO scheme.
	Learning process $LP_i$ is then updated based on the underlying model used, lines~$10-12$. Lines~$14-20$ show a transfer-step. At line~$15$, source is selected among the agents w.r.t. some fixed criteria~$SS$. Secondly, remaining agents, i.e., targets, apply a filtering~$TM$ over source transfer buffer and then sample a batch composed by a fixed number~$B$ of tuples~($17$). Finally, at line~$18$, each target agent updates its learning process based on the obtained batch.

\begin{figure}[htb!]
		\centering
		\includegraphics[width=1.\columnwidth]{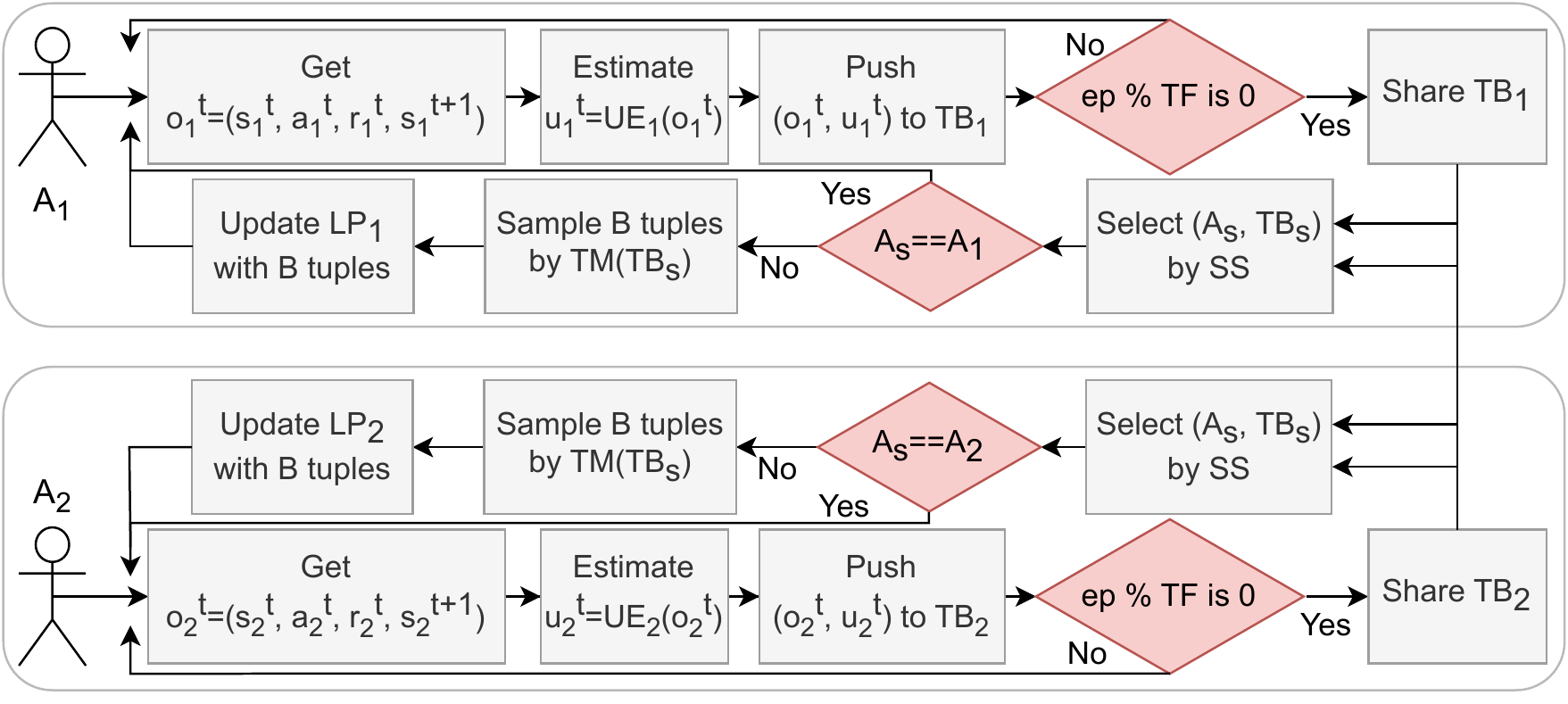}
		\caption{\acrshort{ONEES} workflow with a 2 agents scenario.}
		\label{fig:ONEES_wflow}
	\end{figure}

	While the above provides a full description of the algorithm, Fig.~\ref{fig:ONEES_wflow} sketches \acrshort{ONEES} workflow for the simplest scenario with only 2 agents.

	\subsection{Source Selection Criteria}\label{ss:SS_metrics}
	One of the most crucial parts of \acrshort{ONEES} is accurately selecting the source from which knowledge is extracted. In this section, we propose \textit{Source Selection Metrics~(SS)} to select a source agent among several candidates. 
	Roles are dynamically assigned at the beginning of each transfer iteration and \textit{SS} goal is to identify an experienced candidate whose knowledge will benefit others.

	We propose two different methods to identify transfer source in ~\acrshort{ONEES}: \acrfull{u} and \acrfull{bp}. \acrfull{u} relies on average uncertainty~$\overline{u_i}$ over transfer buffer~$TB_i$.
	Therefore, source agent $A_s \leftarrow min_{i=0}^{N}(\overline{u_i})$ where, $\overline{u_i}~=~\frac{\sum_{t=0}^{|TB_i|} u_i^t \in TB_i}{|TB_i|}$.
	On the other hand, \acrshort{bp} relies on performance achieved by agents during latest episodes. As performance measure we use average cumulated reward overs episodes' finite-horizon undiscounted return~($\overline{R_i}$). 
	$A_s \leftarrow max_{i=0}^{N}(\overline{R_i})$ where, $\overline{R_i}~=~\frac{\sum_{e=0}^{E} \sum_{t=0}^{\tau^e_i} r^{e,t}_i}{E} $. $E$ is the number of evaluated episodes and $\tau^e_i$ represents the length of the $e$\textit{-th} episode for the $i$\textit{-th} agent.
	Hence, average sum of rewards returned by the environment over episodes from initial to goal state with a finite number of steps.

	\subsection{Transfer Filtering Criteria}\label{ss:TM_metrics}
	Another crucial part of \acrshort{ONEES} is accurately selecting the  knowledge to be transferred in each source and target transfer, as shown in line~$17$ of Algorithm~\ref{alg:online_trasfer}. 
	In this section we introduce \textit{Transfer Filtering Criteria~(TM)} to prioritise certain tuples based on their score over a set of measures.

	To simplify the readability of this section we describe the simplest case with two agents, which, at each transfer step, are referred to as source~$A_s$ and target~$A_t$.
	We rely on two criteria to identify experiences to be transferred, \textit{uncertainty} and \textit{expected surprise}.
	\textit{Expected surprise}~\cite{white2014surprise}, is defined over target agent and is approximated through \textit{Temporal Difference-error}~(TD-error). 
	For instance, given a DQN with predictor and target networks, TD-error is defined as Mean Squared Error~(MSE) between the two outputs.
	
	Assuming that $UE_s$ and $UE_t$ use a standard method with fixed parameters, we use their estimated uncertainties to select relevant tuples to be transferred. In detail, given ($o_s^i, u_s^i$), an observation with its epistemic uncertainty explored by~$A_s$~at~time~$i$,
	then $A_t$ estimates current uncertainty~$u_t^{o_s^i}$ over the interaction~$o_s^i$ sampled by $A_s$. Hence,  $u_t^{o_s^i} = UE_t(o_s^i)$.
	We then define $\Delta-$\textit{conf} as discrepancy between the two estimations: 
	$\Delta-$\textit{conf}$=u_t^{o_s^i} - u_s^i$ .
	Note that $u_t^{o_s^i}$ is an epistemic estimation at the time of transfer and changes over~time.

	As a result, $A_t$ receives a personalised batch of experience that aims to fill shortcomings in its belief.
	Based on these two criteria we can define multiple filtering functions for incoming knowledge.
	In this paper, we introduced the following criteria to prioritise transferred experience:
	
	\begin{itemize}
		\item{\acrshort{rdc} $-$ \acrlong{rdc}.} By choosing this technique, an agent randomly samples $B$ interactions with an associated $\Delta-$\textit{conf} above the median computed across all samples.
		\item{\acrshort{hdc} $-$ {\acrlong{hdc}}. Agent selects the top $B$ entries with higher $\Delta$\textit{-conf}.}
		\item{{\acrshort{lec}} $-$ {\acrlong{lec}}. While previous filters are defined only over uncertainty, this one also considers expected surprise. To balance the different scales, values are normalised within a $[0,1]$ interval and then summed up. Finally, target agent pulls $B$ tuples from the buffer with the highest values.} 	
	\end{itemize}

	The criteria presented above are explicitly or implicitly used in other TL frameworks. Nonetheless, we are the first to explicitly apply them to online transfer learning based on experience sharing, and evaluate \acrshort{ONEES} performance while exploring different combinations of \textit{SS} and \textit{TM} to assess their impact.

	\section{Evaluation Setup}\label{sec:simulations}

	We demonstrate \acrshort{ONEES} by using two different off-policy Deep-RL methods: the Dueling DQN~\cite{wang2016dueling} in Cart-Pole and \acrshort{pp} and Deep Deterministic Policy Gradient with parametrised action space~(PA-DDPG)~\cite{Hausknecht_ddpg2015} for \acrshort{HFO}. 

	\begin{table}
		\renewcommand*{\arraystretch}{1.1}
		\caption{\label{tab:ONEES_PARAM} \acrlong{ONEES} parameters.}
		\centering
		\begin{threeparttable}
			\begin{tabular}{cccc}
				\hline
				\multicolumn{1}{c}{\textbf{Parameter}}&\textbf{Cart-Pole} &\textbf{\acrshort{pp}}&\textbf{\acrshort{HFO}}\\\hline
				\multicolumn{1}{c}{{N}}
				& 5 & 4  & 3 \\			
				\multicolumn{1}{c}{{TF}}
				& 200 & 300 & 400 \\			
				\multicolumn{1}{c}{{TB Capacity}}
				& $10,000$ & $100,000$ & $25,000$ \\			
				\multicolumn{1}{c}{{\textit{SS}:\acrshort{bp} Eval. Interval}}
				& $200$ & $400$ & \xmark\\			
				\multicolumn{1}{c}{{Ep. Start Transfer}}
				& $600$ & $2,500$ & $2,400$\\			
				\multicolumn{1}{c}{{Max Timestep}}
				& $400$ &$200$ & $500$\\
				\multicolumn{1}{c}{{Max Episode}}
				& $1,800$ &$8,000$ & $20,000$\\ \hline			
			\end{tabular}
		\end{threeparttable}
	\end{table}
	
	Table~\ref{tab:ONEES_PARAM} presents the parameters used to assess \acrshort{ONEES} across the evaluated tasks. In addition, for Cart-Pole and \acrshort{pp}  we also vary $B:\{500,1500,5000\}$, $SS:\{\overline{U}, BP\}$, and $TM:\{\textrm{{\acrshort{rdc}}},\textrm{{\acrshort{hdc}}},\textrm{{\acrshort{lec}}}\}$, obtaining 18 transfer settings defined on $B\times SS \times TM$.
	In \acrshort{HFO}, we have constrained the evaluation to a single transfer setting which has proved the best trade-off between positive transfer and cost of transfer over simpler environments, with the following parameters: $B=100$, $SS=\overline{U}$ and $TM=\textrm{{\acrshort{hdc}}}$.
	
	While \acrshort{pp} and \acrshort{HFO} are naturally multi-agent environments, for Cart-Pole we used multiple parallel and independent instances. We collected $20$ runs for Cart-Pole and \acrshort{pp} to study the impact of different transfer settings and $10$ runs for \acrshort{HFO}.

	We compare \acrshort{ONEES} against \textit{no-transfer}, where transfer is disabled, and two different advice-based baselines with and without external expertise:

	\begin{itemize}
		\item \acrshort{noexp_adv} $-$ {\acrlong{noexp_adv}}, based on~\cite{ilhan2019teaching}, to evaluate the impact of sharing action-advice versus experience.
		For each state visited by an agent $A_i$, uncertainties are estimated by all agents and whether $A_i$ is the most uncertain then it asks for advice. Other agents provide their best estimated action and $A_i$ uses majority voting to take the final action. Allocated budget matches the number of our total interaction transferred per agent with $B=5,000$, hence, $30,000$ for Cart-Pole and $85,000$ for \acrshort{pp}.
		\item  \acrshort{exp_adv} $-$ {\acrlong{exp_adv}} to benchmark our expert-free transfer framework against the state-of-the-art expert-based teacher-student framework. We provide a jury, composed by $3$ trained agents, to ward off any bias induced by using a single trained agent as teacher. An agent asks for an advice whenever its uncertainty is higher than a threshold and advice is selected as the most frequent action returned from the jury. Advice is then given until episode terminates.
	\end{itemize}

	We implemented~\acrshort{exp_adv} with $5$ heads on advantage branch for dueling DQN. Each head provides $|A|$ estimation of advantages and we approximate uncertainty by normalising head's logit within a $[0,1]$ interval, as such, uncertainty ranges between $[0, 0.3]$.  We set the threshold for Cart-Pole to $0.25$ and allocated a budget of $300$ episodes. For \acrshort{pp}, after several trials, threshold is set to $0.02$ and maximum budget to $6,000$ episodes. 
	For consistency with \acrshort{noexp_adv}, in Section~\ref{sec:results} we report used budget in terms of overall action-advice suggested.

\begin{figure}[htb!]
		\centering
		\includegraphics[width=.57\columnwidth]{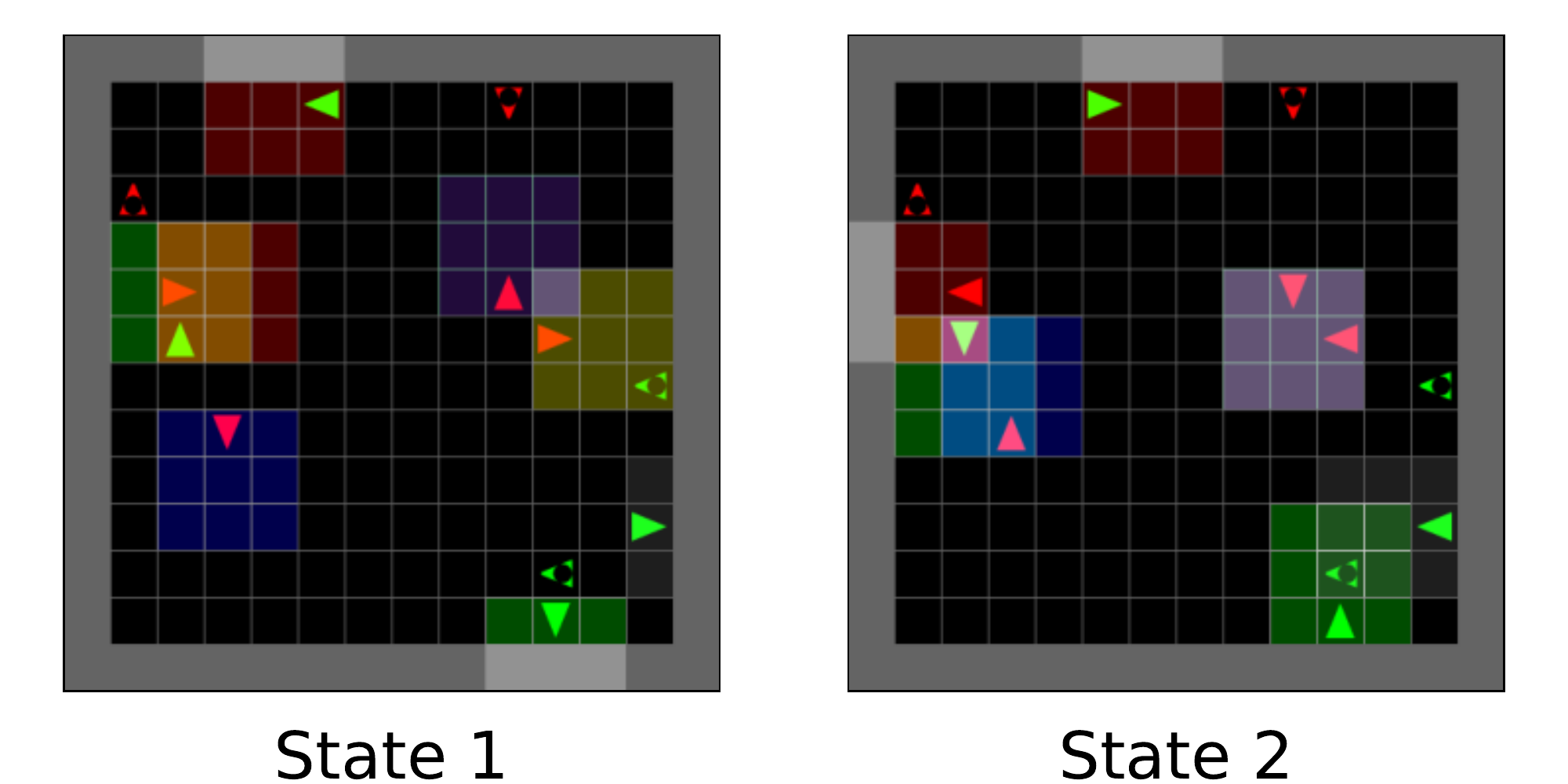}
		\caption{\acrfull{pp} environment. $8$ predators (filled oriented triangles), and $4$ prey (dotted oriented triangles), are fairly split into two colour-based teams. 
		}
		\label{fig:pp_states_rotaiton}
	\end{figure}

	For {\acrshort{pp}}, different versions exist and for our  configuration we rely on a grid-based implementation~\cite{Castagna2022MultiAgentTL}. A depiction of this problem is reported in Figure~\ref{fig:pp_states_rotaiton}.
	$8$ predators and $4$ prey are fairly divided into two colour-based teams and randomly spread in a $12\times12$ grid.

	We control each predator with a RL agent while prey follow a random policy. An episode terminates when one or both teams have no more prey left to capture. Each agent perceives a $3\times3$ grid centered around the first cell it is facing, as highlighted in Figure~\ref{fig:pp_states_rotaiton}.
	Thus, observation is composed of a $3$-dimensional $3\times3$ matrix. First channel describes object type, i.e., \textit{void}, \textit{wall}, \textit{predator}, \textit{prey}. Second channel identifies the team-membership, i.e., \textit{none} when not applicable, \textit{red} and \textit{green}. Third channel provides rotation of an object, i.e., \textit{none} when non applicable, \textit{up}, \textit{down}, \textit{left}, \textit{right}. Predators have $5$ possible actions: \textit{rotate left}, \textit{rotate right}, \textit{forward}, \textit{pick} and \textit{hold}. 
	First two actions rotates a predator left or right while remaining in the same cell. \textit{Forward} moves an agent to the next cell it is facing, if it is empty. When facing a prey, \textit{pick} is used to catch the prey. 

	Finally, while all predators are controlled by RL agents, only a single team is enabled to share knowledge, in order to better isolate performance impact of sharing. Results are therefore presented as aggregated performance across the \textit{sharing-enabled} team.

	\begin{table}
		\renewcommand*{\arraystretch}{1.1}
		\caption{\label{tab:DDQN_params} Dueling Q-Network parameters.}
		\centering
		\begin{threeparttable}
			\begin{tabular}{r cc}
				\hline
				\multicolumn{1}{c}{\textbf{{Parameter}}}& \textbf{{Cart-Pole}} & \textbf{{\acrshort{pp}}} \\\hline
				
				\multicolumn{1}{c}{{Input Layer}}
				& FC($4,128$)& Conv1d($3, 7, k=1$)\\
				
				\multicolumn{1}{c}{\multirow{1}{*}{Hidden}}
				& \multirow{2}{*}{FC($128,64$)}& Conv1d($7, 15, k=1$)\\
				\multicolumn{1}{c}{\multirow{1}{*}{Layer(s)}}&&FC($135,256$)\\
				
				\multicolumn{1}{c}{{Value Branch}}
				& FC($64,1$)& FC($256, 1$)\\
				
				\multicolumn{1}{c}{{Adv. Branch}}
				& FC($64,2$)& FC($256, 5$)\\
				
				\multicolumn{1}{c}{{Activation}}
				& ReLU& ReLU\\
				
				\multicolumn{1}{c}{{Loss}}
				& MSE& MSE\\
				
				\multicolumn{1}{c}{{Optimiser}}
				& Adam & Adam \\
				
				\multicolumn{1}{c}{{Learning Rate}}
				& $1e-5$ & $1e-5$\\
				
				\multicolumn{1}{c}{{Target Replacement}}
				& $1,000$ & $10,000$\\
				
				\multicolumn{1}{c}{{Exploration}}
				& Softmax & $\epsilon$-decay\\
				
				\multicolumn{1}{c}{{Replay Buffer Size}}
				& $10,000$ & $10,000$\\
				
				\multicolumn{1}{c}{{Batch Size}}
				& $32$ & $32$\\
				
				\multicolumn{1}{c}{{Ep. Start Training}}
				& $100$ & $100$\\\hline
			\end{tabular}
		\end{threeparttable}
	\end{table}
	
	Dueling DQN architectures are specified in Table~\ref{tab:DDQN_params} while approximation of Q-values~(Q) is given by combining the two output streams of the network as follow $Q = value + adv-\overline{adv}$. 
	In Cart-Pole we balance exploration and exploitation through \textit{Softmax} function~\cite{sutton1998introduction} while in {\acrshort{pp}} we follow an $\epsilon$-decay probability that eventually anneals to $0$ at episode $7,450$.
	Model is optimised at each time-step after the $100^{th}$ episode.
	As replay buffer, we used the proportional version of prioritised experience replay with suggested parameters~\cite{schaul2015prioritized}.

	 \begin{figure}[htb!]
	 	\centering
	 	\includegraphics[width=.52\columnwidth]{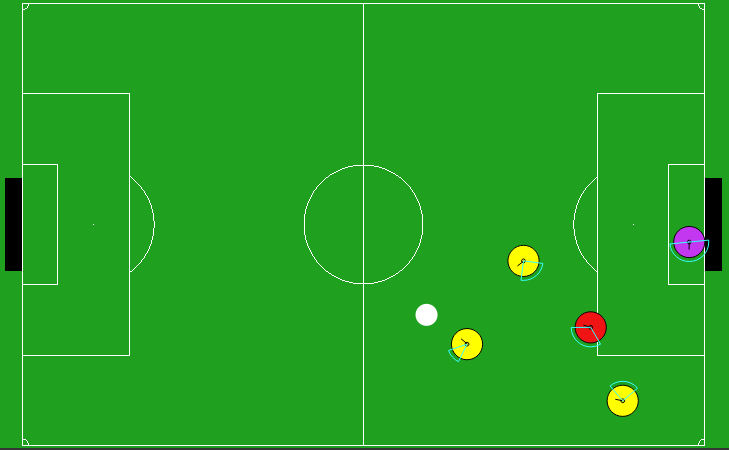}
	 	\caption{Instance of \acrlong{HFO} 3v2. Offense is depicted in yellow, defense in red and goalie in violet.}
	 	\label{fig:hfo_example}
	 \end{figure}
 
 	Lastly, we compare \acrshort{ONEES} against a no-transfer baseline and \acrshort{exp_adv} in \acrshort{HFO}. We omitted \acrshort{noexp_adv} as synchronizing all the player agents, in a complex scenario, for each time-step to evaluate their confidence is unfeasible. Our \acrshort{HFO} setup is a 3v2 where offense players are controlled by our agents with noise free perception and defense by \textit{HELIOS} baseline~\cite{ALA16-hausknecht}. Figure~\ref{fig:hfo_example} depicts a snapshot of our configuration.

	For \acrshort{exp_adv}, we empirically found it more advantageous to provide single-step advice rather than full trajectory needed to accomplish an episode. Thus, we allocated a maximum budget of $10,000$ per agent and we fixed the uncertainty threshold to $1e-4$.

	As PA-DDPG architecture for \acrshort{HFO}, we use an existing configuration publicly available~\cite{pddpg-hfo}.

	\section{Results}\label{sec:results}

	This section presents the results and analysis of ~\acrshort{ONEES} performance. As a preliminary study, we also first evaluate the performance of our proposed \acrshort{sarnd} uncertainty estimator against RND, to compare the sensitiveness of the two models.
	
	\subsection{\acrshort{sarnd} Evaluation}
	To assess \acrshort{sarnd} capability we use the \acrshort{pp} benchmark. Each estimator visits 2 states depicted by Fig.~\ref{fig:pp_states_rotaiton} while sampling different actions. We fixed architectures of both target and predictor across the two estimators, but their input layer dimensions differ as \acrshort{sarnd} requires extra neurons to accommodate action, reward and next state.  
	Agents perform action $a_0$ for $250$ steps to strengthen the prediction of both uncertainty estimators. Afterward, action sampled is changed to~$a_1$ for $25$~steps. Finally, agents resample~$a_0$ for~$25$ steps.
	
	\begin{figure}[!hb]
		\centering
		\includegraphics[width=.74\columnwidth]{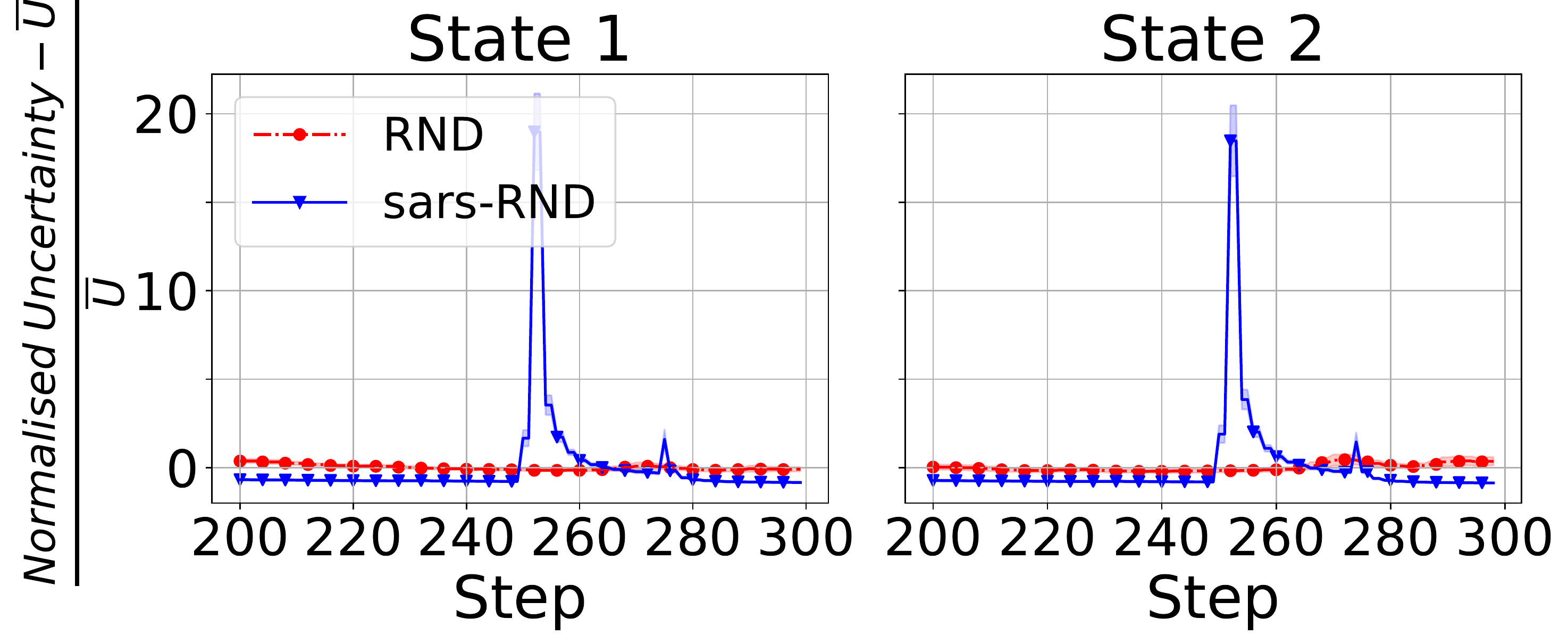}
		\caption{RND and \acrshort{sarnd} normalised estimated uncertainty with their 95\% confidence interval over states depicted by Figure~\ref{fig:pp_states_rotaiton} with different actions sampled over time.
		}
		\label{fig:zoom_uncertainty_comparison}
	\end{figure}
	
	Overall, we observed higher uncertainty within the first steps while curves decrease asymptotically to $0$ for both estimators as they increasingly visit the states. However, by evaluating uncertainty in a narrowed interval as in Figure~\ref{fig:zoom_uncertainty_comparison}, we notice that while RND keeps a flat trend, \acrshort{sarnd} registers a spike in uncertainty on the change of action, firstly, at step~$250$ when action $a_0$ is replaced by $a_1$, and secondly, at~$275$ when $a_0$ is resampled. Second spike is smaller compared to first as \acrshort{sarnd} recognises the previous seen interaction.

This confirms thats \acrshort{sarnd} provides a more reliable uncertainty estimation by considering executed action alongside visited state while RND is not sensitive to action variation. Furthermore, given the observed decreasing trend, \acrshort{sarnd} generalises well enough to recognise familiar states even while sampling a different action. Thus, uncertainty estimated for previously un-encountered state is higher than previously seen state while exploring different actions.

	\subsection{\acrshort{ONEES} Evaluation}


	\paragraph*{Cart-Pole results} Cart-Pole is a simple environment with a continuous observation space and a binary decision for the control policy. We evaluate the performance of our TL technique by evaluating the learning curve over the episodes.
	
	Given the minimalistic environment, generally we have observed very similar performance across all the techniques: \textit{no-transfer}, action-advice based baseline and ~\acrshort{ONEES}.
	\begin{figure}[!htb]
		\centering
		\begin{subfigure}[b]{\columnwidth}
			\centering
			\includegraphics[width=.85\columnwidth]{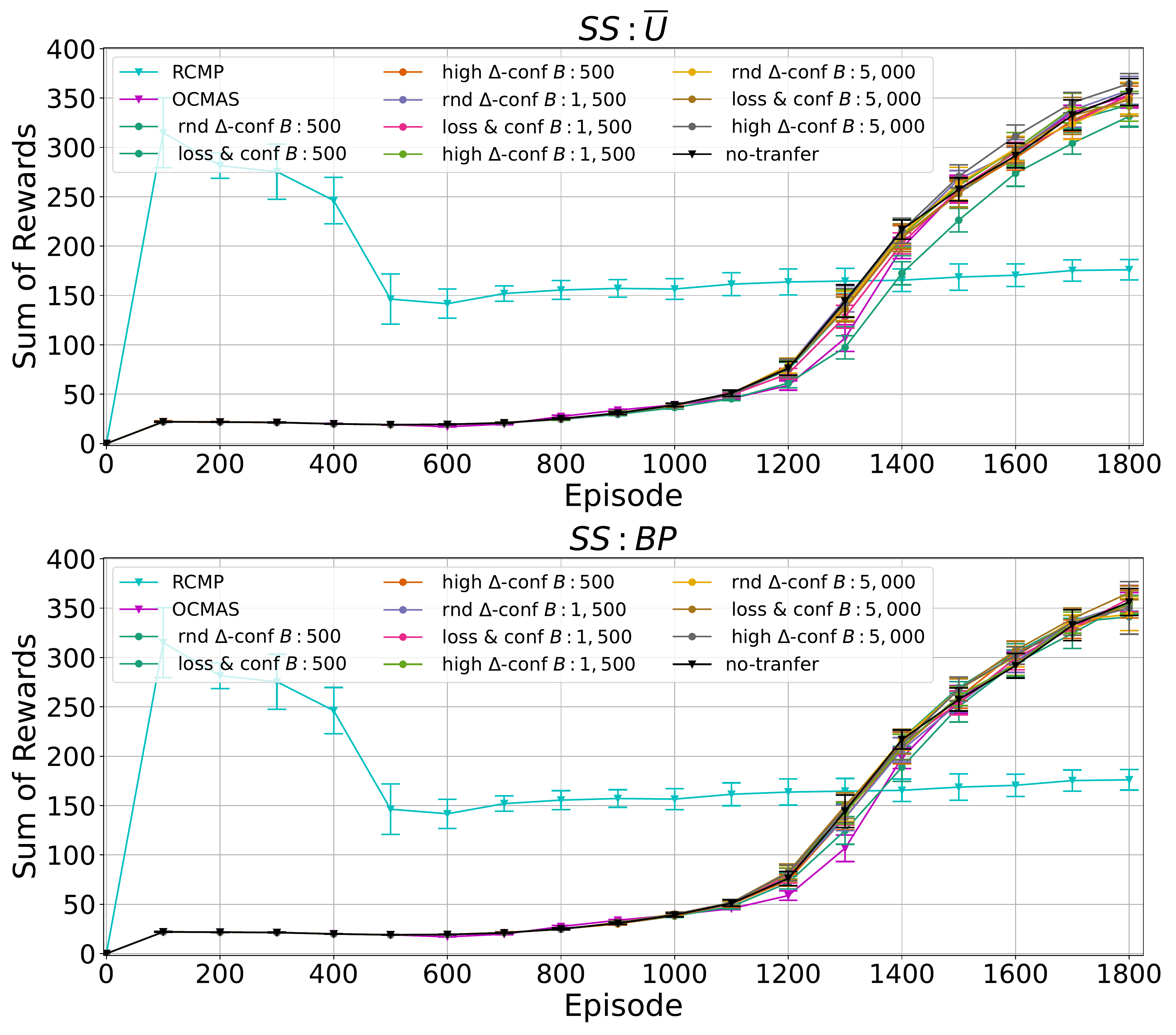}
		\end{subfigure}
		\begin{subfigure}[b]{\columnwidth}
			\centering
			\includegraphics[width=.7\columnwidth]{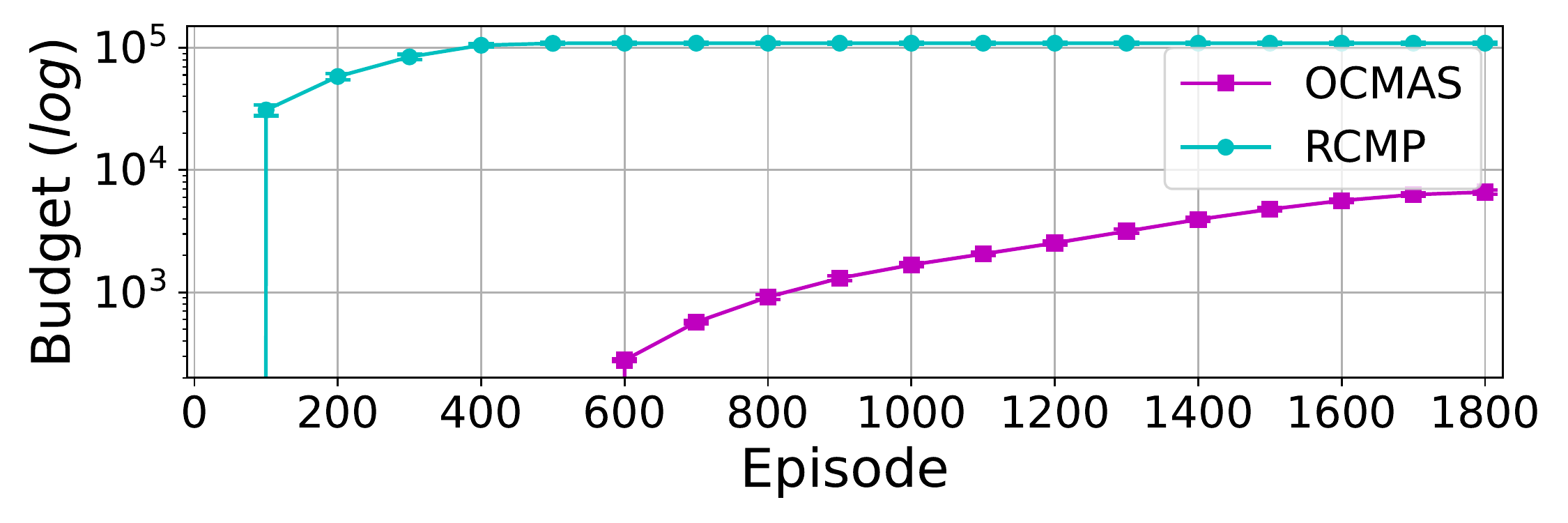}
		\end{subfigure}
		\caption{Learning curves in Cart-Pole and budget used by advice-based baselines.}
		\label{fig:cp_results}
	\end{figure}
	Figure~\ref{fig:cp_results} shows the learning curves with their 95\% confidence interval. 

	We observe a fluctuating transfer impact for~\acrshort{ONEES} achieving both positive and negative transfer.
	\acrshort{ONEES} positive transfer starts around episode $1,400$, and its benefit is generally linear to quantity of shared information. Despite the mild improvement,
	we cannot establish a configuration that prevails overall but we observe that different transfer settings impact differently the performance throughout agent's life. We conclude that in this scenario, \acrshort{ONEES} has room for improvement by enabling dynamic selection of techniques used to select both the source and experience to transfer.

	Based on our evaluation, \acrshort{exp_adv} shows jump-start within first episodes~($0-1,200$) while later performance is kept to a similar value until the end. Thus, despite the initial improvement, at the end it caps the target's performance. 
	
	
	\paragraph*{\acrlong{pp} results} While in Cart-Pole we observed low improvements by our approach, \acrshort{pp} shows a more interesting outcome. We measured the performance of a sharing-enabled team in \acrshort{pp} through \textit{Average Reward}, \textit{Average Catch} and \textit{Win Probability}.
	We do not report the distribution of miscaught prey since all \acrshort{ONEES} agents learn not to catch other team's prey.
	
\begin{figure}[!htb]
		\centering
		\begin{subfigure}[b]{\columnwidth}
			\centering
			\includegraphics[width=.73\columnwidth]{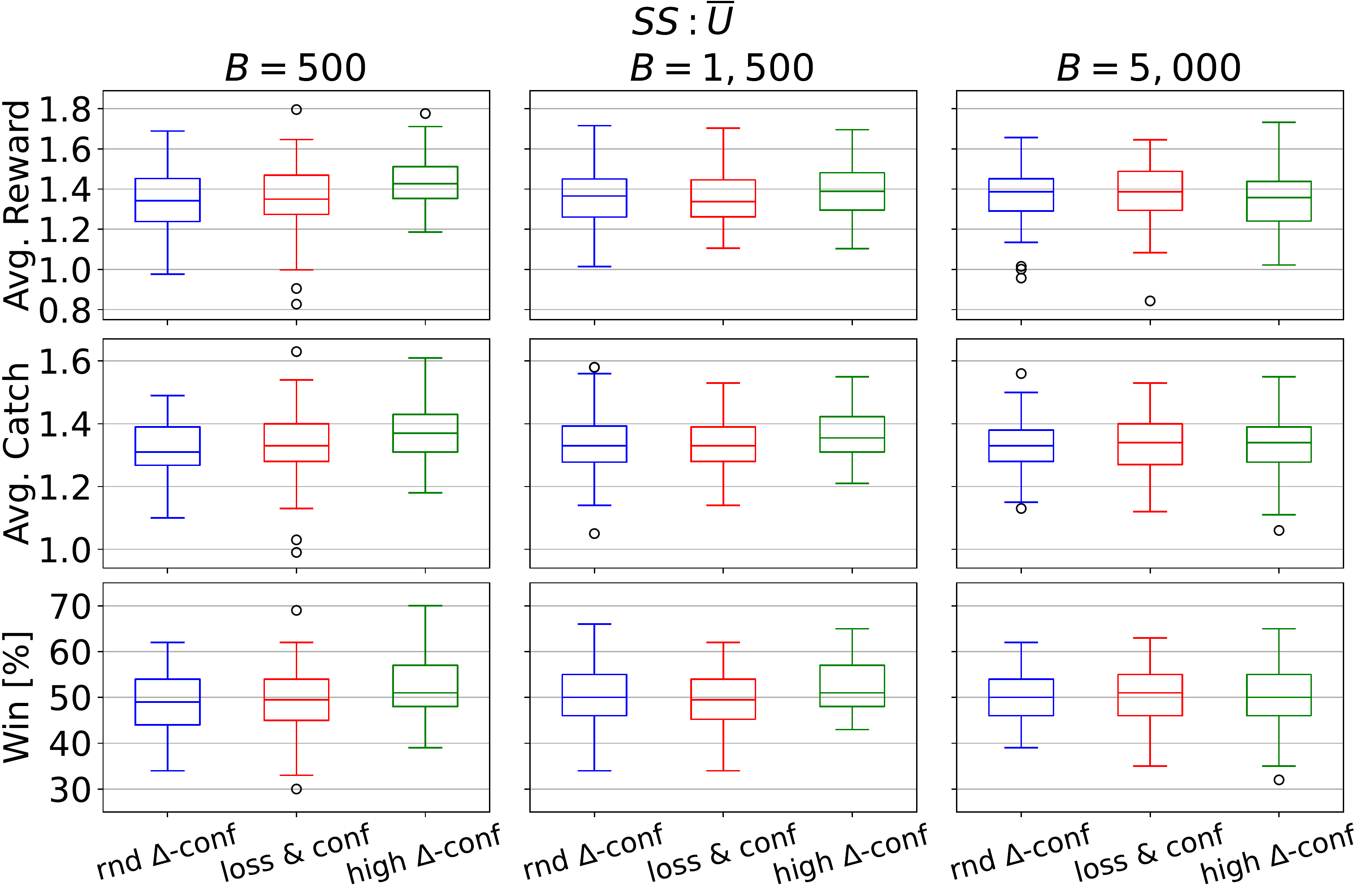}
		\end{subfigure}
		\begin{subfigure}[b]{\columnwidth}
			\centering
			\includegraphics[width=.73\columnwidth]{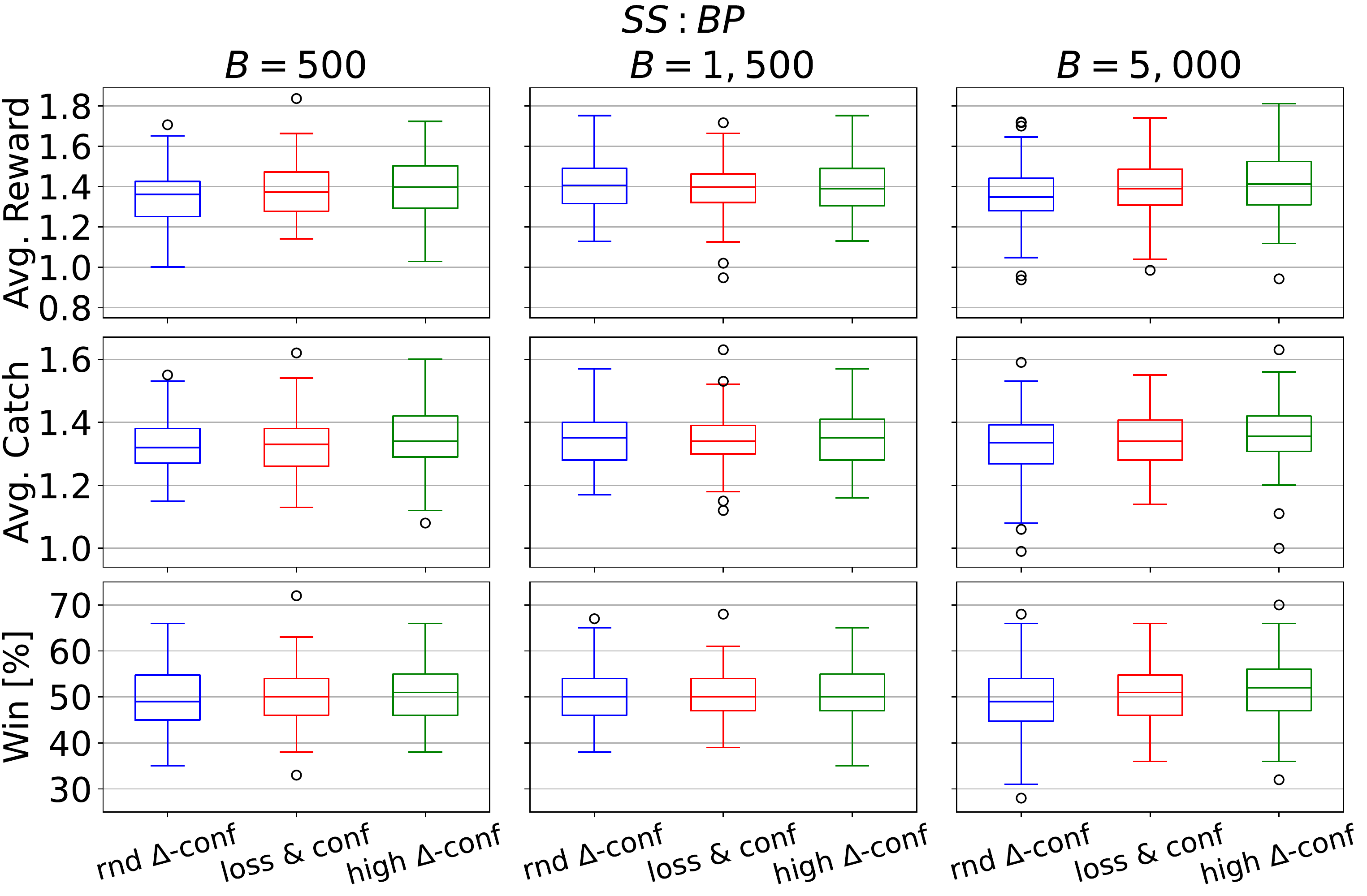}
		\end{subfigure}
		\caption{Performance achieved by \acrshort{ONEES} transfer-enabled team in \acrshort{pp} across 500 test episodes.}
		\label{fig:mtpp_EF_OnTL_results}
	\end{figure}

	First, we report performance across the $18$ transfer settings defined for~\acrshort{ONEES} in Figure~\ref{fig:mtpp_EF_OnTL_results}. 
	All predators learnt a similar behaviour to fulfil their goal. Generally, all variations have a common median value across all performance metrics and the difference between one \acrshort{ONEES} configuration to the other is very small. This common trend is opposed to what we previously observed in Cart-Pole where the benefit of transfer was generally correlated to budget used. This could suggest that for more challenging environment a lower budget may be the best trade-off to keep low transfer overhead cost and to improve agent's performance.

	Furthermore, while fixing budget $B$ and $SS$, generally using \acrshort{rdc} performs the worst, \acrshort{hdc} performs the best, while \acrshort{lec} usually lies in between. Thus, overall from start of sharing to end, filtering experience to share by discrepancy between source and target uncertainty (i.e., \acrshort{hdc}), is the most effective in this type of environment.

\begin{figure}[!htb]
		\centering
		\begin{subfigure}[b]{\columnwidth}
			\centering
			\includegraphics[width=.81\columnwidth]{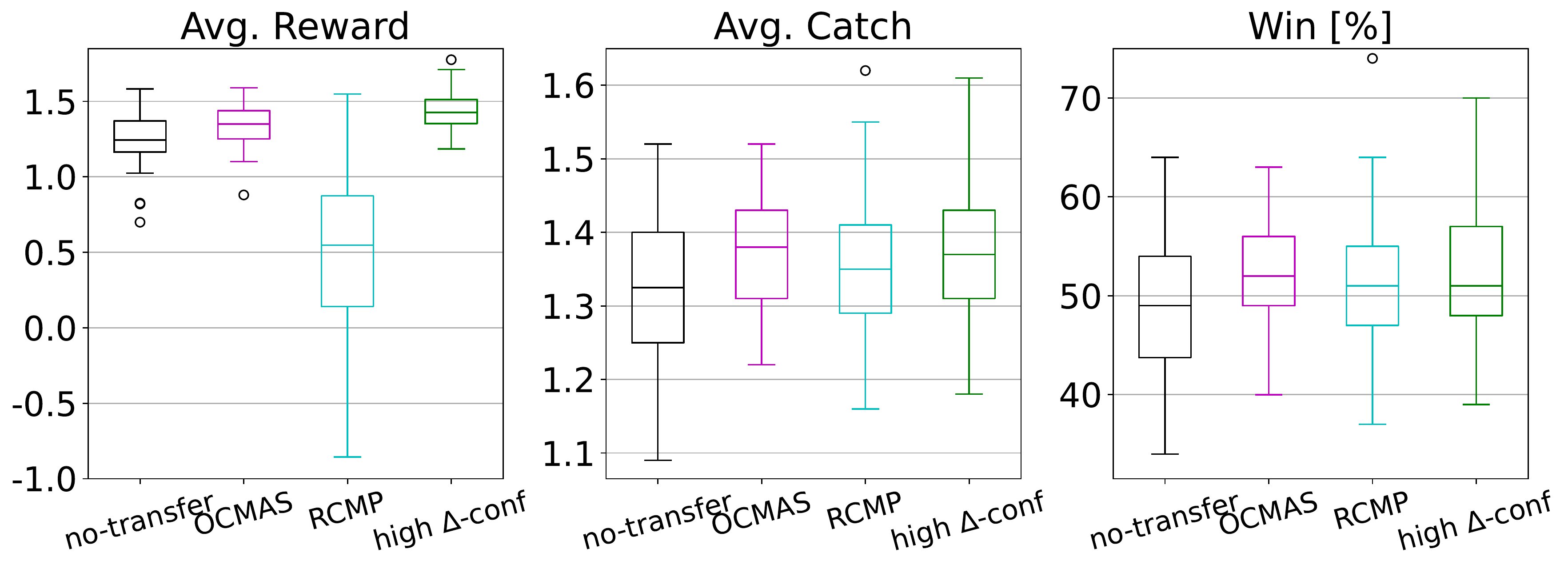}
		\end{subfigure}
		\begin{subfigure}[b]{\columnwidth}
			\centering
			\includegraphics[width=.7\columnwidth]{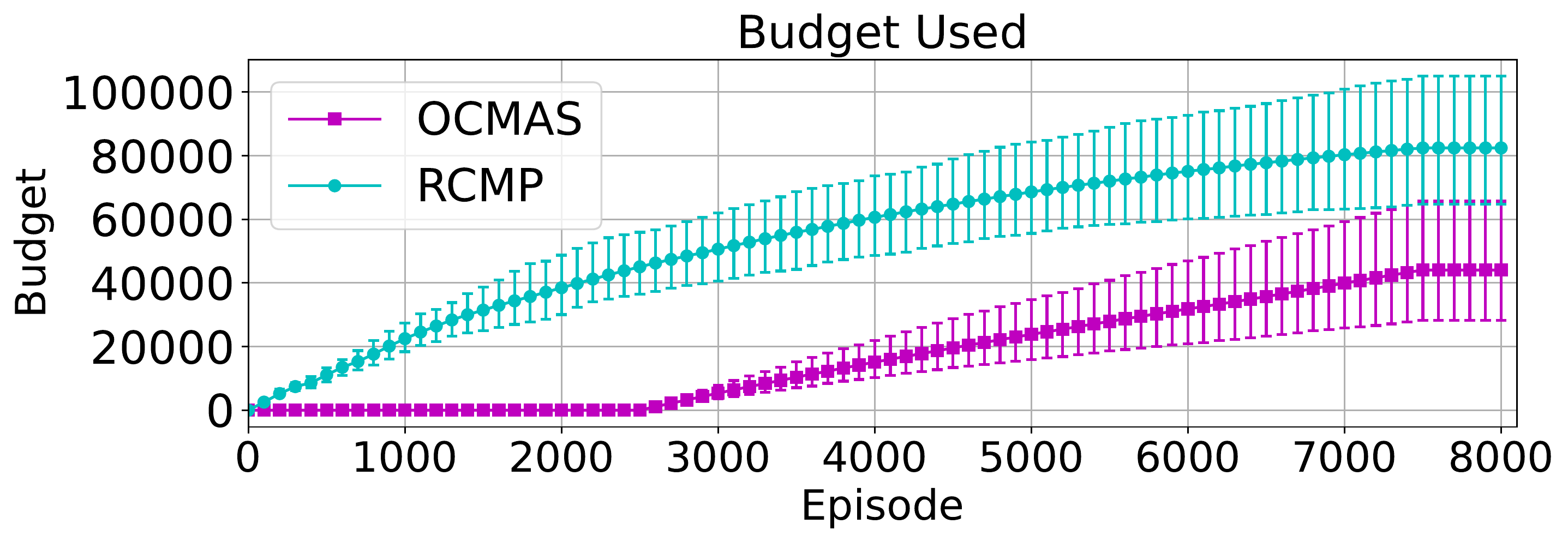}
		\end{subfigure}
		\caption{Team based \acrshort{ONEES} performance in \acrshort{pp} against the baselines with budget consumption.}
		\label{fig:mtpp_res_baselines}
	\end{figure}

	Figure~\ref{fig:mtpp_res_baselines} shows the comparison between \textit{no-transfer},   \acrshort{noexp_adv} and \acrshort{exp_adv} baselines, and the best \acrshort{ONEES} configuration: $B=500$, $SS=\overline{U}$ and $TM=$ \acrshort{hdc}.
	Contrary to what we expected, \acrshort{exp_adv} does not significantly outperform the others and actually decreases the average reward. With \acrshort{exp_adv} predator has learnt to perform more frequently expensive actions, i.e., catch. Despite this different behaviour leads to similar win and positive catch distributions, overtime leads also to lose reward and to miscatch preys.
	
	\paragraph*{\acrlong{HFO} results} \acrshort{HFO} is a multi-agent environment where agents need to collaborate to score a goal. Collaborative interaction complicates the task that already has sparse reward function and continuous state and control space.	
	Thus, in a 3 vs 2 situation, having a single poor performing agent within the team will likely result in capping the team performance.
	
\begin{figure}[!htb]
		\centering
		\includegraphics[width=.83\columnwidth]{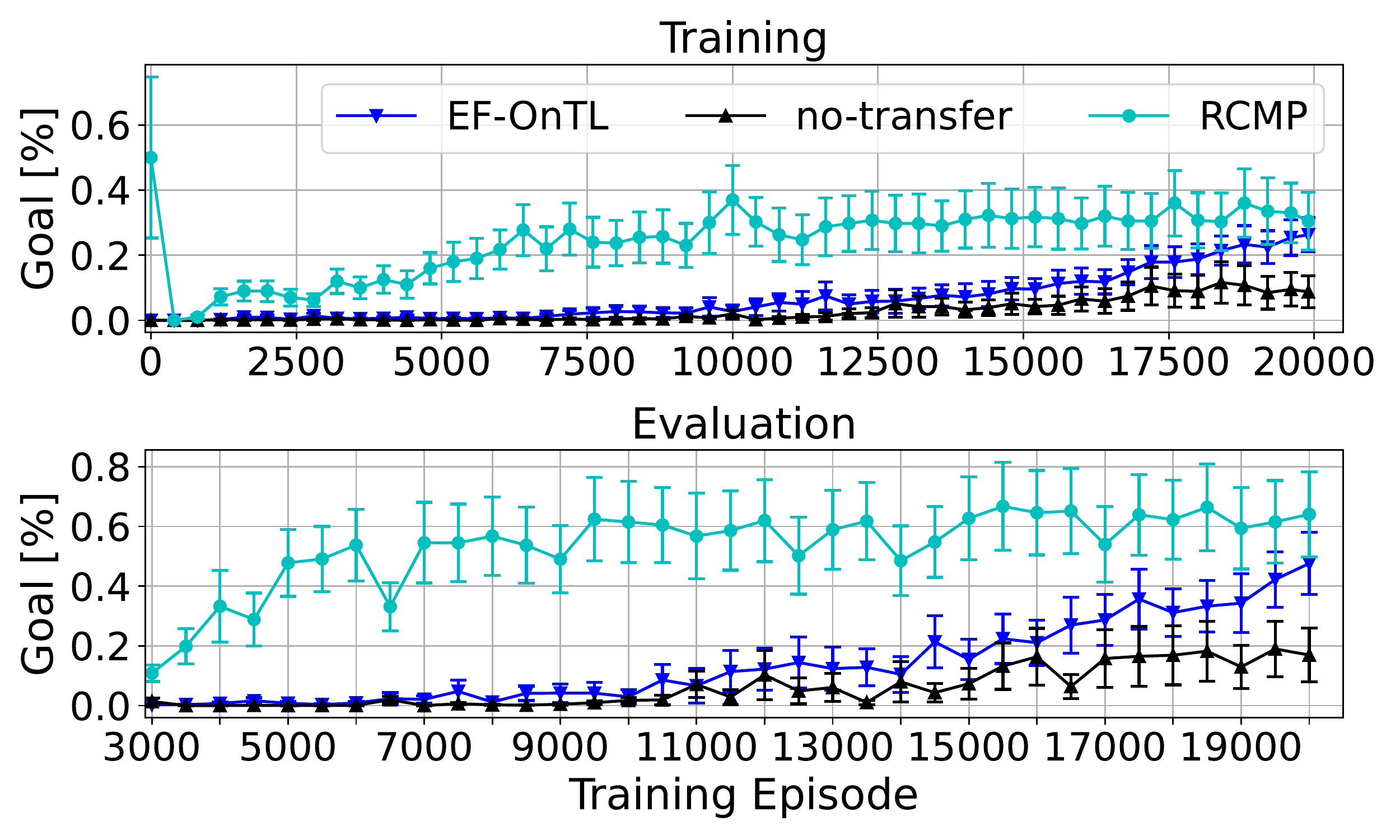}
		\caption{ \acrshort{HFO} goal probability during training and testing.		
		}
		\label{fig:HFO_results}
	\end{figure}

	To evaluate \acrshort{ONEES} against \textit{no-transfer} and \acrshort{exp_adv} baseline, we measure probability of the team scoring a goal during training and testing phases. We estimate probabilities by counting the number of accomplished episodes within a sample. For training we use a sliding window of $100$ episodes while on testing we run $500$ episodes where agents use their up-to-date knowledge. These results are reported by Figure~\ref{fig:HFO_results}.

	\acrshort{ONEES} shows good performance in both training and testing when compared against \textit{no-transfer} and their difference increases over time. Within the latest $5,000$ episodes, \acrshort{ONEES} powered agents, on average, doubled their goal probability when compared against \textit{no-transfer} agents.

	In \acrshort{HFO} the exploitation of expert knowledge plays a crucial role. In fact, \acrshort{exp_adv} achieves outstanding results since an early stage of training. However, despite an initial jump-start, \acrshort{exp_adv} goal probability seems to fluctuate within the same range from episode $7,000$ onwards resembling Cart-Pole trend.

	\paragraph{\acrshort{exp_adv} additional notes}While implementing \acrshort{exp_adv} as baseline, we encountered some challenges to be addressed that are worth to be mentioned. 
	Firstly, in Cart-Pole, given the normalisation to maintain a fixed scale for uncertainty and binary action decision, estimator shows low granularity in uncertainty estimation. In fact, uncertainty falls into 3 possibilities based on agreement across the $5$ heads, $0$, when all agree, $0.2$ when a single head disagree, and $0.3$ when two heads disagree with others.
	Secondly, in environment with multiple actions as \acrshort{pp}, the algorithm performance are very sensitive to the chosen threshold.
	Despite a lower threshold leads agent to be guided for longer, the final performance in learning curve and tracked metrics are significantly lower. We provide a deeper analysis on \acrshort{exp_adv} baseline in supplementary material.
	Finally, in \acrshort{HFO} expert teachers significantly improve student's performance. Although, in PA-DDPG, the ensemble is on critic which outputs a single value and as such is impossible to predict a range and hence to normalise the estimated uncertainty. Consequently, the uncertainty curve decreases sharply within the first hundreds episodes and then ranges within a narrow interval making it again not trivial to find the right threshold given the tight uncertainty interval.

	To conclude, while \acrshort{exp_adv} requires careful tuning on threshold, our proposed \acrshort{ONEES} does not require tuning or adaptation other than budget~$B$ and transfer frequency~$TF$. Furthermore, \acrshort{ONEES} has shown comparable performance with both~\acrshort{noexp_adv} and \acrshort{exp_adv} in Cart-Pole and \acrshort{pp} while significantly lowering the communication and synchronisation cost by only sharing every $N$ episodes a personalised batch of experience tailored to target agent gaps.

	In \acrshort{HFO}, the use of \acrshort{ONEES} led the offense team to score more than twice as many goals as the no-transfer agents. Furthermore, in the latest $5,000$ episodes, the gap between \acrshort{exp_adv} and our novel \acrlong{ONEES} faded away over time.

	\section{Conclusion}\label{sec:conclusion}

	This paper introduced \acrfull{ONEES}, a novel dynamic online transfer learning framework based on experience sharing and suitable for multi-agent implementation, and \acrfull{sarnd}, an extension of RND, to estimate agent epistemic uncertainty from a full RL interaction. \acrshort{ONEES} is evaluated varying budget used, teacher selection criteria and experience filtering criteria to improve shared batch quality across multiple agents.

	
	We benchmark \acrshort{ONEES} against no-transfer, \acrshort{exp_adv} and \acrshort{noexp_adv} in three environments, Cart-Pole, \acrlong{pp} and \acrlong{HFO}. 
	
	\acrshort{ONEES} has shown a significant improvement when compared against no-transfer baseline and overall similar, and in some conditions superior, performance when compared against advice-based baselines. Despite the minor improvement in performance, the communication cost is significantly reduced by sharing less frequently an experience batch from a temporary selected teacher to a target agent. The transferred batch is personalised upon the expertise of source and target shortcomings.
	Finally, this paper open up several directions for future work. 
	Firstly, in our experiments on~\acrshort{ONEES} we did not cover any empirical evaluation on transfer frequency nor optimal batch-size. Instead, we fixed this values to some pre-defined parameters as we preferred to study source selection and transfer filtering criteria, presented respectively in Section~\ref{ss:SS_metrics} and Section~\ref{ss:TM_metrics}, used to design the transferred batch. 
	Furthermore, experiments in Cart-Pole have highlighted how aforementioned criteria and budget affect TL performance throughout different stage of learning. 
	This requires further study to assess \acrshort{ONEES}~ influence over target while enabling dynamic tuning of these parameters.
	Secondly, despite~\acrshort{sarnd} can be dropped to minimize used resources when there is no-need of uncertainty, i.e., on deployment, \acrshort{sarnd}~could be optimised by limiting the input to state action pairs rather than full agent-environment interaction. The proof of equivalence while reducing the input is yet to be provided.
	Thirdly, we assumed that agents are homogeneously defined and all are engaged during a transfer step. Thus, their task is equally defined in observation space, action space, goal, reward model and transition function. This paper does not cover adaption of \acrshort{ONEES} and \acrshort{sarnd} across different tasks or goals.
	Lastly, uncertainty-labelled experience could be stored and used later to train a novice agent in an offline scenario.

	\ack This work was sponsored, in part, by the Science 	Foundation Ireland under Grant No. 18/CRT/6223 	(Centre for Research Training in Artificial Intelligence), and SFI Frontiers for the Future project 21/FFP-A/8957.

	\bibliography{ecai}
\end{document}